\documentclass[conference]{IEEEtran}
\IEEEoverridecommandlockouts
\usepackage{cite}
\usepackage{amsmath,amssymb,amsfonts}
\usepackage{algorithmic}
\usepackage{graphicx}
\usepackage{textcomp}
\usepackage{xcolor}
\usepackage{float}
\usepackage{graphicx}
\usepackage{booktabs,ragged2e} 
\def\BibTeX{{\rm B\kern-.05em{\sc i\kern-.025em b}\kern-.08em
    T\kern-.1667em\lower.7ex\hbox{E}\kern-.125emX}}
\begin{document}

\title{A Novel Deep Learning Method for Segmenting the Left Ventricle in Cardiac Cine MRI
}

\author{\IEEEauthorblockN{\textsuperscript{} Wenhui Chu $^{\dag}$,  Aobo Jin $^{\dag}$, Hardik A. Gohel $^{\dag}$}\\

\IEEEauthorblockA{$^{\dag}$\textit{Dept. of Computer Science, University of Houston-Victoria,  Victoria, USA.} \\ChuW1@uhv.edu,  Jina@uhv.edu, gohelh@uhv.edu}

}

\maketitle

%\author{\IEEEauthorblockN{\textsuperscript{} Wenhui Chu ^{\dag}$,   Nikolaos V. Tsekos ^{\dag}$}\\

%\author{\IEEEauthorblockN{\textsuperscript{} Wenhui Chu $^{\dag}$, Giovanni Molina $^{\dag}$, Nikhil V. Navkar $^{\dag}^{\dag}$, Christoph F. Eick $^{\S}$, }{Aaron T. Becker $^{\P}$, Panagiotis Tsiamyrtzis $^{\ast}$, Nikolaos V. Tsekos $^{\dag}$}\\

%\IEEEauthorblockA{$^{\dag}$\textit{MRI Lab, Dept. of Computer Science, University of Houston, Houston, USA.} \\wchu@uh.edu, gemolinaramos@uh.edu, nvtsekos@central.uh.edu}
%\IEEEauthorblockA{$^{\dag}^{\dag}$\textit{Dept. of Surgery, Hamad Medical Corporation, Doha, Qatar.} \\nnavkar@hamad.qa}
%\IEEEauthorblockA{$^{\S}$\textit{ DAIS Lab, Dept. of Computer Science, University of Houston, Houston, USA.} \\ceick@uh.edu
%}
%\IEEEauthorblockA{$^{\P}$\textit{Dept. of Electrical and Computer
%Engineering, University of Houston, Houston, USA. } \\atbecker@uh.edu}
%\IEEEauthorblockA{$^{\ast}$\textit{Dept. of Statistics, Athens University of Economics and Business, Greece.} \\
%pt@aueb.gr}

%}

%\maketitle

\begin{abstract}

This research aims to develop a novel deep learning network, GBU-Net, utilizing a group-batch-normalized U-Net framework, specifically designed for the precise semantic segmentation of the left ventricle in short-axis cine MRI scans. The methodology includes a down-sampling pathway for feature extraction and an up-sampling pathway for detail restoration, enhanced for medical imaging. Key modifications include techniques for better contextual understanding crucial in cardiac MRI segmentation. The dataset consists of 805 left ventricular MRI scans from 45 patients, with comparative analysis using established metrics such as the dice coefficient and mean perpendicular distance. GBU-Net significantly improves the accuracy of left ventricle segmentation in cine MRI scans. Its innovative design outperforms existing methods in tests, surpassing standard metrics like the dice coefficient and mean perpendicular distance. The approach is unique in its ability to capture contextual information, often missed in traditional CNN-based segmentation. An ensemble of the GBU-Net attains a 97\% dice score on the SunnyBrook testing dataset. GBU-Net offers enhanced precision and contextual understanding in left ventricle segmentation for surgical robotics and medical analysis.

\end{abstract}


\section{Introduction}
As reported by the World Health Organization in 2011, cardiac diseases are the cause of 17.5 million deaths worldwide every year [1]. U-Net has become noteworthy for its precise segmentation capabilities, which are essential for determining clinical boundaries. Additionally, it makes efficient use of GPU memory, which is important because GPU memory can often be a bottleneck when it comes to computing power. U-Net distinguishes itself by achieving high-precision segmentation with fewer annotated training examples, offering a substantial benefit given the high cost of acquiring these samples. Cardiac MRI imaging is crucial for diagnosis, yet its manual analysis is laborious and costly, requiring expert input and not always yielding error-free results. With the expansion of computational power, neural networks are increasingly being employed across various aspects of life, especially in medical imaging for the classification, segmentation, and detecting objects. There is a general agreement that neural networks improve the efficiency and accuracy of image processing. Specifically, in the context of left ventricle imaging, better segmentation precision has the potential to transform the diagnosis of cardiac diseases, which could lead to improved survival rates in the future. As accurate segmentation is crucial for assessing cardiac functions, there has been a significant increase in research on left ventricle imaging in the past years [2]. This paper focuses on the enhancement of U-Net by exploring diverse normalization strategies and contrasting their effects.

Substantial advancements have been achieved recently in the application of neural networks for segmenting images. For instance, Melkemi \textit{et al.} [3] introduced a distributed algorithm for image segmentation that includes multiple segmentation agents and a coordinating agent. Each segmentation agent applies the iterated conditional modes method to produce a near-optimal segmented image, enhancing the precision and effectiveness of the segmentation despite the method's substantial time complexity. Chen \textit{et al.}  [4] devised a model known as artificial co-evolving tribes, which they employed to address the challenge of image segmentation. In this model, individuals within the tribes work, communicating from one agent to another to enhance the uniformity across the collective image regions. Huang \textit{et al.} [5] introduced an innovative segmentation algorithm inspired by artificial ant colonies (AC), drawing parallels between ant self-organization and human cardiac neurons. In their algorithm, each ant retains a memory of a reference object that gets updated upon locating a new target.

Cardiac segmentation has been approached through a purely convolutional neural network [6], while another strategy fused level set methods with deep belief networks for left ventricle delineation [7]. A pivotal ingredient for deep learning's success is an ample dataset, essential for honing accuracy [8]. Prior to these advancements, methodologies like the Imagenet [9], multiple layers of representation [10], handwritten digit recognition[11],  Faster R-cnn [12] and Deep visual-semantic alignments [13] were prevalent. 

Hence, the identification of suitable tools for the evaluation and comparison of segmentation algorithms is crucial. In this paper, we analyze and detail these tools to support developers in enhancing and assessing their algorithms. Our research is centered on implementing various normalization methods within the U-Net structure to enhance the segmentation of left ventricle images. The layout of the paper is structured as such: the methodology section starts by explaining our choice to use U-Net and offers a comparative analysis of the normalization strategies under study, such as batch normalization, group normalization, layer normalization, and batch instance normalization. This section will lead into an explanation of the neural network architectures and the dataset driving our study. Following that, in section III, we will detail the approaches to data preprocessing and the experimental results. The paper will then conclude with a section dedicated to discussions and final remarks.

\section{METHODOLOGY}

\subsection{The Structure of Networks }

U-net represents a convolutional neural network (CNN) that is purpose-built for the segmentation of biomedical images [14]. The U-net framework, with its distinctive U-shaped structure and skip-connection feature, is adept at conserving GPU memory. This is particularly advantageous since GPU memory can often be a restrictive factor in comparison to computational capabilities, making memory efficiency vital for CNN performance. Evolving from U-Net, several derivatives have surfaced, such as DenseNet [15], UNet++ [16], Attention UNet [17], BNU-Net [18], GridNet [19], Mask-RCNN [20] and IBU-Net[21].

In this research, our emphasis is on normalization techniques: batch normalization [22], layer normalization [23],  instance normalization [24], and group normalization[25]. Essentially, these approaches are seen as methods of data optimization. Within deep learning, normalization plays a crucial role in improving the accuracy and speed of training. Especially in cases where features vary greatly, normalization proves to be effective by standardizing these features to a uniform range. This alignment is instrumental in mitigating the internal covariate drift, ensuring stability in network parameters throughout the training phase. Normalization proves to be particularly effective in scenarios where there is significant variability in the distribution of features. By limiting the range of the weights and preventing them from becoming excessively large, normalization can facilitate a more rapid and efficient optimization process.

Our study starts with an analysis of batch normalization [22]. This approach equalizes the output from previous activation stages by normalizing it across each batch in the system. It acts as a systematic fine-tuning of data through each layer of the neural network. In our use case, a batch norm layer is incorporated into the U-Net, which is represented in Figure 1.

\begin{figure}[ht!]
\centering
\includegraphics[width=0.45\textwidth]{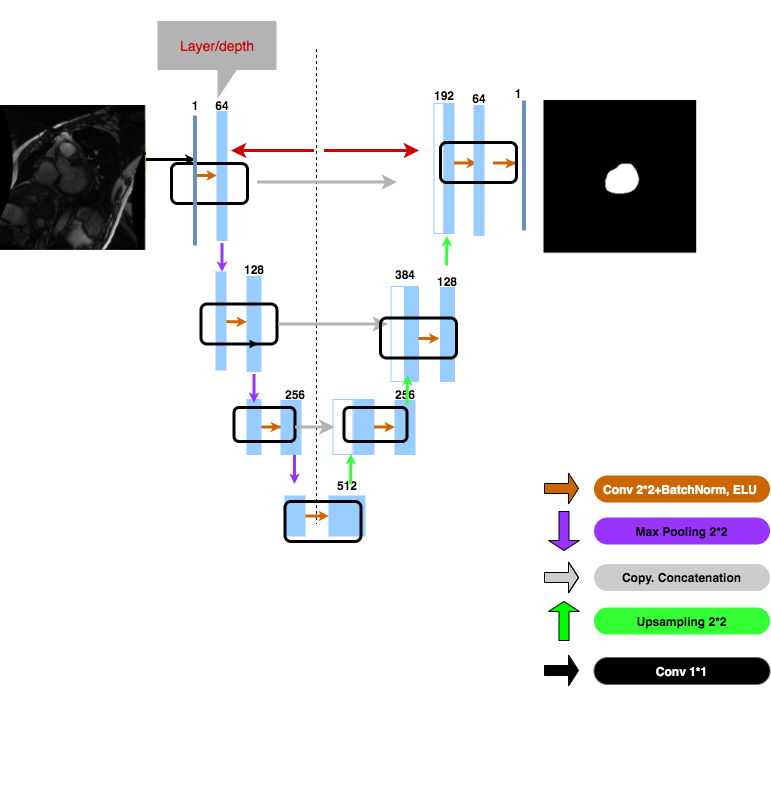}
\caption{
\label{fig:BNUarch}
The BNU-Net convolutional network is architecturally composed of two primary components: (a) the contraction path designed for feature extraction, and (b) the systematic application of batch normalization following each convolutional operation within the convolutional layers.}
\end{figure}

Batch Normalization aims to mitigate internal covariate shift, significantly speeding up the training process of deep neural networks [22]. In such settings, normalization is applied vertically to counter the issues arising from redundant rescaling.

Despite its benefits, batch normalization presents certain limitations. It does not represent the preeminent approach for mitigating internal covariate shift within sophisticated neural network structures [22]. During the normalization of outputs from preceding layers, the method involves dividing the batch by its standard deviation and then modifying it with the computed mean. This approach, though, tends to underperform in the context of online learning, resulting in reduced generalization capabilities. This shortfall is primarily attributed to the variability in batch sizes across successive iterations, which adversely affects the consistency and reliability of the learning process. Such alterations in input data can hamper the overall efficacy. In response to these challenges, alternative normalization strategies have emerged.

Conversely, layer normalization [23] operates by standardizing inputs across different features, representing a shift away from the batch-focused methodology employed by batch normalization. The calculation of mean and variance in layer normalization is analogous to that in batch normalization, where for each data point, the mean and variance are ascertained on a per-feature basis, see Figure 2. Group Normalization (GN) is presented as a straightforward alternative to Batch Normalization (BN). GN partitions the channels into distinct groups and determines the mean and variance within each group for the purpose of normalization. The computational process of GN is not contingent on batch sizes, and it maintains consistent accuracy across a broad spectrum of batch sizes[25]. Our methodology incorporates group normalization within each convolutional segment, with the underlying foundation being the exponential linear unit (ELU). A visual representation is available in Figure 3. 
%, where i stands for the batch and j stands for the feature. You can see the difference between batch normalization and layer normalization in Figure 3 [18]. 
\begin{figure}[ht]
\centering
\includegraphics[width=0.45\textwidth]{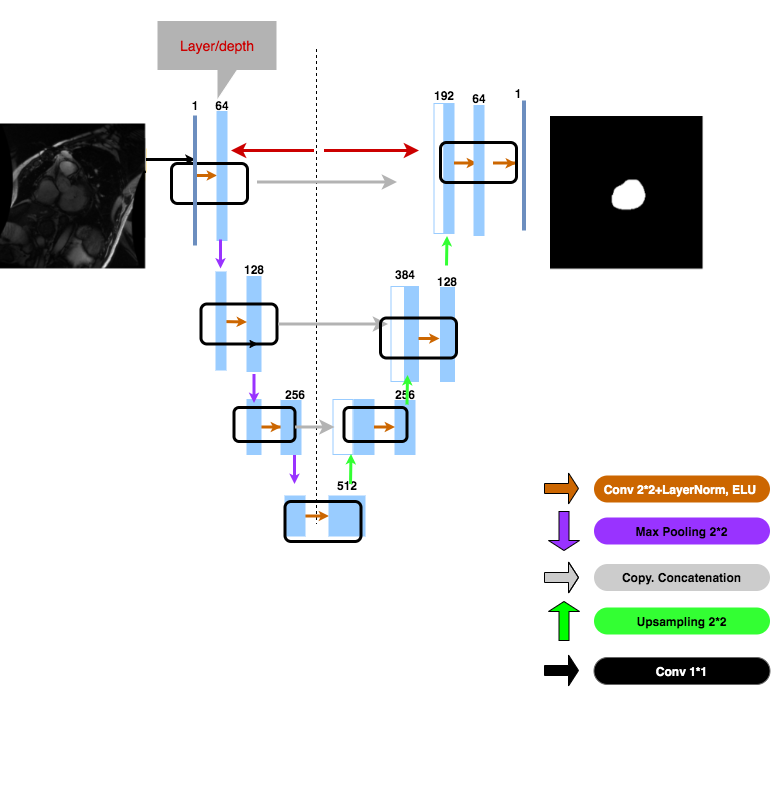}
\caption{
\label{fig:LNUarch}
The architecture of the proposed fully convolutional network, designated as LNU-Net, integrates layer normalization processes following each convolutional operation within the convolutional layers.}
\end{figure}

\begin{figure}[ht]
\centering
\includegraphics[width=0.45\textwidth]{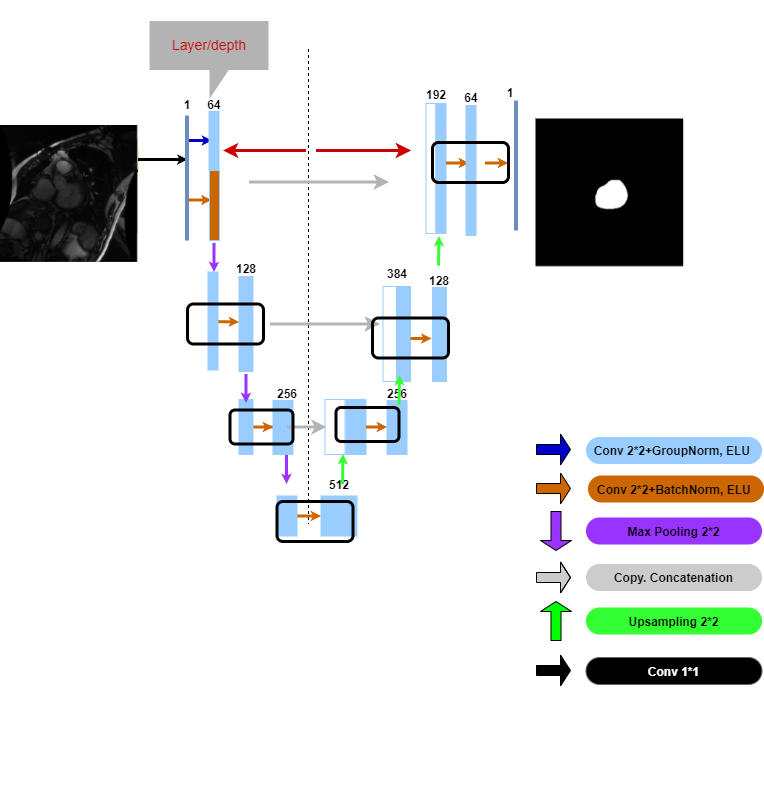}
\caption{
\label{fig:GBUarch}
The GBU-Net, a fully convolutional network, is characterized by its unique architecture. It systematically incorporates both group normalization and batch normalization after each convolution within its convolutional layers.}
\end{figure}

Batch normalization normalizes the features using the mean and variance calculated within a mini-batch, contrasting group normalization, where  doesn't depend on the batch size, making it particularly useful for tasks where the batch size is small due to memory constraints. The study titled ``Group Normalization" [25] affirmed that group normalization is recognized as an effective normalization layer that does not utilize the batch dimension.

The batch group normalization method effectively combines the principles of both batch normalization and group normalization. This innovative approach unifies the strengths of both techniques to enhance network performance. Batch normalizes the feature map using the mean and variance determined across the batch, height, and width dimensions of a feature map, subsequently re-scaling and shifting the normalized feature map to preserve the representational capacity of Deep Convolutional Neural Networks (DCNN). However, this technique is not devoid of limitations. Notably, during the inference phase, the applicability of batch-wise normalization is not feasible. Consequently, the mean and variance are predetermined, typically through a cumulative average, based on the training dataset [26]. This predetermined approach obviates the need for normalization during the testing phase. It is important to acknowledge that these pre-calculated statistics are subject to potential fluctuations in response to alterations in the target data distribution. This methodology, while effective, is susceptible to fluctuations that can introduce inconsistencies during various stages such as training, transfer, and testing. A particularly notable concern is that a reduction in batch size can significantly impact the accuracy of batch statistics estimation, as indicated in [25]. To mitigate these issues, group normalization employs a hyper-parameter, designated as G, which regulates the number of feature instances for statistical calculation. This ensures the provision of consistent statistical data, unaffected by varying batch sizes and devoid of erratic or ambiguous values.

To further address these challenges, our approach integrates both group and batch normalization within the initial convolutional block, incorporating an Exponential Linear Unit (ELU) and forwarding the resultant output to subsequent layers, as depicted in Figure 3. This strategy effectively harmonizes the strengths of both normalization methods, allowing the model to adeptly adjust their blending ratios through gradient descent. As corroborated by prior research [27], the concept of batch group normalization (BGN) has been proposed. This technique aims to rectify the inaccuracies in statistical calculations by Batch Normalization (BN) at either small or extremely large batch sizes. The introduction of channel, height, and width dimensions in BGN acts as compensatory measures, enhancing the overall robustness and accuracy of the normalization process.

\begin{figure}[ht!]
\centering
\includegraphics[width=0.45\textwidth]{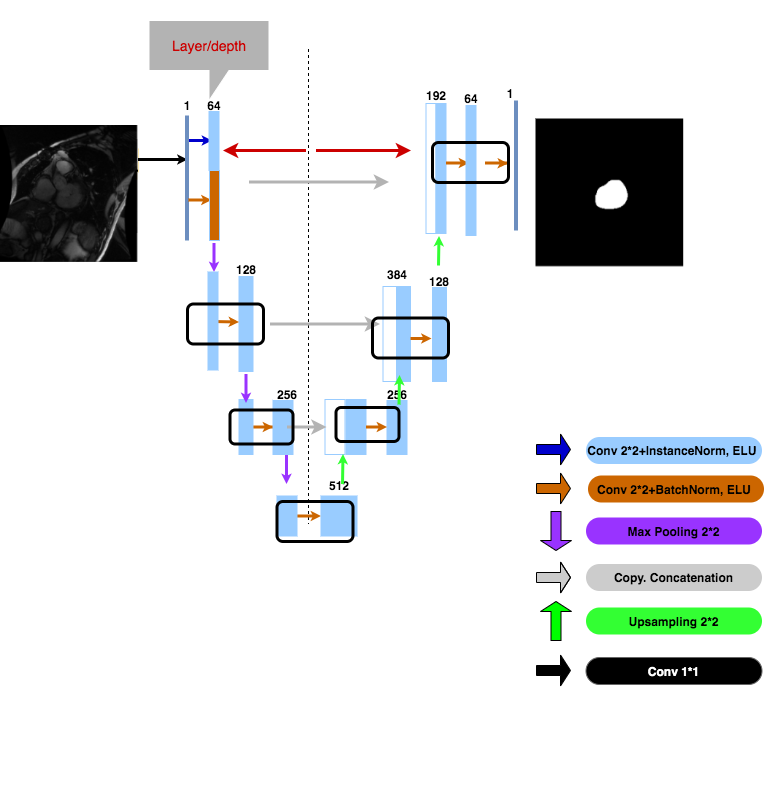}
\caption{
\label{fig:IBUarch}
Architecture of the IBU-Net convolutional network.  Instance normalization is applied in the first convolutional layer. Batch normalization is performed after each convolution in the convolutional layer.}
\end{figure}

\begin{table}
  \centering
  \caption{COMPARISON BETWEEN 
BNU-NET MODEL WITH ELU AND BNU-NET WITH ReLU}\label{tab:elu-relu-compare}
   \begin{tabular}{cp{13mm}p{13mm}p{13mm}}\toprule  &BNU-Net with ELU  &BNU-Net with ReLU   \\\midrule
    Dice mean &\textbf{0.92}  & 0.90\\
    Dice std & \textbf{0.04} &0.06\\
    Sensitivity & \textbf{0.96} &0.95 \\

    \bottomrule
  \end{tabular}
\end{table}

\subsection{Dataset}

For our research purposes, we employed the Sunnybrook dataset [28], sourced from the Imaging Research  for Cardiovascular Intervention at Sunnybrook Health Sciences Centre. Crucially, the dataset was specifically curated to target left ventricular structures for the MICCAI 2009 Left Ventricle Segmentation Challenge. The MRI images are encoded in the DICOM (Digital Imaging and Communications in Medicine) format, which includes a range of metadata parameters pertaining to both the patient and the imaging specifics. For each patient record, a corresponding set of manually delineated contours is provided. These ground truth contours, drawn by expert professionals, are available in text files comprising the contour points. The dataset employed in this study comprises a total of 805 images, which have been sourced from 45 patient cases. For the purposes of analysis, this dataset was methodically partitioned into training, validation, and testing subsets, adhering to an equitable distribution ratio of 15:15:15.  All tests were performed using an NVIDIA GeForce Titan X Pascal GPU. While assessing the computational speed per epoch, it's essential to factor in the performance of the computational resources. The dataset encompasses a diverse array of cardiac conditions, including 12 cases each of heart failure with infarction and without infarction, hypertrophy, along with 9 cases involving healthy patients. The number of images per patient varied, ranging from 12 to 28.  Each sequence within the dataset comprises 6 to 12 two-dimensional cine stacks, characterized by a slice thickness of 8 millimeters. Additionally, these cine stacks feature an in-plane resolution ranging from 1.3mm to 1.4mm.

\subsection{Core Model}

For image segmentation, our approach involved using a revised U-Net model, which is built upon a fully convolutional network framework. We enhanced this U-Net variant by integrating a variety of techniques to improve its efficiency in cardiac image segmentation. Specifically, Exponential Linear Units (ELU) were used as the activation function instead of the ReLU found in the standard version, and a normalization layer was added, which was not present in the original U-Net design. This modification was made to better comprehend the architecture of the model, refer to Figures 1-4.

- Conv2*2 denotes a convolution layer featuring a 2*2 kernel.

- Normalization: This ensures feature map normalization to tackle the internal covariate shift.

- We opted for ELU over ReLU to expedite training and mitigate the vanishing gradient issue, as ELU provides a non-zero value for negative x inputs.

- Cropping2D functions as a layer to trim feature maps. It's also employed to minimize concatenation and subsequently reduce the risk of overfitting.

- Concatenating fuses diverse feature maps from the downsampling process.

- The UpSampling layer enlarges the feature map size.

Our adapted U-Net, designed for the automated segmentation of the left ventricle, is based on the core architecture of a BNU-Net neural network framework [18]. A significant improvement was made to the encoder component. By using an optimized encoder configuration, we managed to reduce the computational burden, thereby making the training process more efficient, particularly within the limitations of the Titan X's processing capabilities.

Our focus on normalization forms a crucial part of our methodology. We experimented with four different normalization techniques – batch normalization, layer normalization, batch instance normalization, and batch group normalization – to fine-tune our model. In our experimentation with U-Net for medical image segmentation, the issue of internal covariate shift posed a significant challenge. Recognizing that training is not dependent on the original data distribution, we found that incorporating data normalization was an effective solution. Additionally, our attention turned to the skip connections in U-Net. We observed that the segmentations from U-Net often suffered from a lack of clarity at the edges, leading to fuzzy boundaries between different areas of the images, which is an inherent issue in the original U-Net design.

\section{RESULTS}

\subsection{Experimental Studies}

The development of efficient deep learning models necessitates a simultaneous reduction in both training and validation errors. Data augmentation is a powerful strategy to achieve this goal. By expanding the dataset, the model encounters a wider spectrum of potential data scenarios, which helps in minimizing the gap between training and validation errors, and subsequently with any test sets [26]. Therefore, data augmentation techniques are employed to improve the quality of medical image datasets. The concept of elastic deformation, introduced by Patrice and colleagues [29] in 2003, is one such technique. We apply elastic deformation to our training images, effectively expanding the dataset size and enhancing the model adaptability. In this study, our data augmentation approach included various techniques such as affine transformations, elastic deformations, and rotations, thereby broadening the dataset diversity and adaptability, as illustrated in Figure 5.

%(Figure 4).  %In order to expand the training data set, We rotate the %training images from 10 degree to 90 degree(Figure 5).

%\begin{figure}[ht]
%\centering
%\includegraphics[width = 8.7cm]{auge.png}
%\caption{ Performing the preprocessing
%and data-augmentation steps.
%(a) Sunnybrook Image and corresponding label. (b) Perform affine transformations (translation=0.03, rotation=[-4.6, 4.6] and scaling=[0.98,1.02]) (c) Perform elastic transformation to augment the training set, in which alpha=[28, 30], sigma=[3.5, 4.0] (Alpha: float for fixed value or [lower, upper] for random value from uniform distribution. Sigma: float, sigma of gaussian filter that smooths the displacement fields.) (d) Rotate -10 degree of the original image.
%}
%\label{fig:preprocessing}
%\end{figure}

\begin{figure*}[htbp]
\centering
\includegraphics[width=7.3in,height=4.3in]{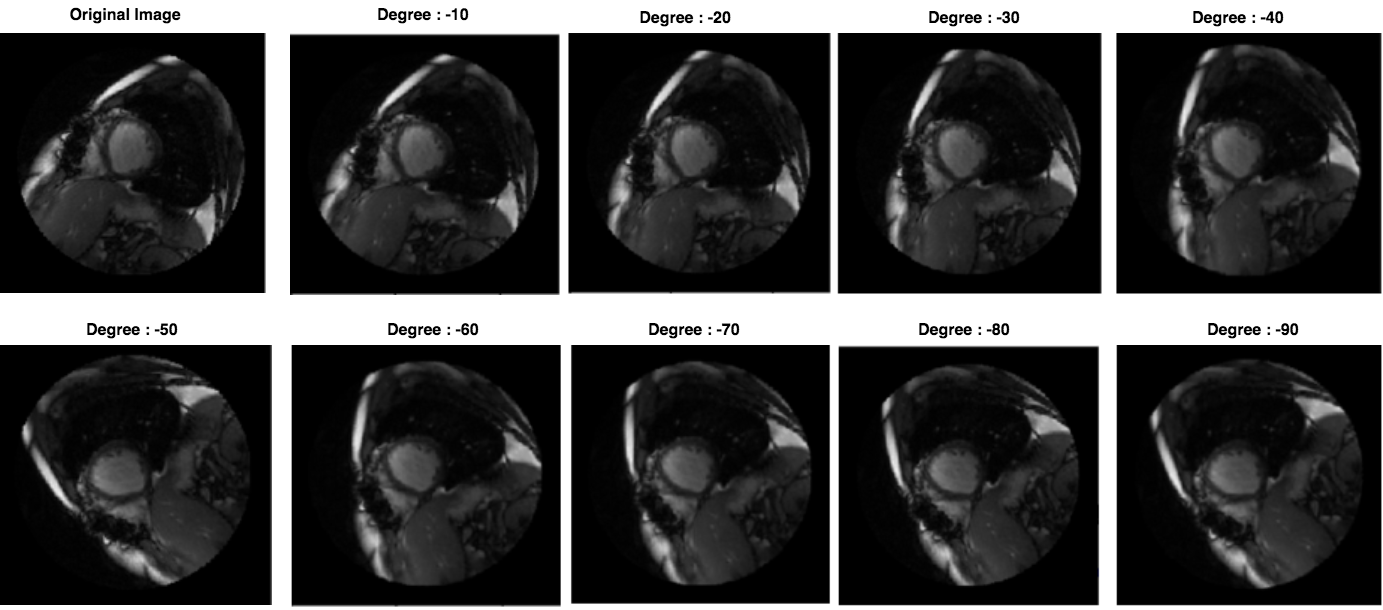}
\caption{Overview of data rotation steps in the proposed sampling method}
\label{fig:rotation}
\end{figure*}

The traditional U-Net model is frequently applied in the segmentation of left ventricle cardiac images. To evaluate its effectiveness, we initially tested it on an NVIDIA GeForce Titan X Pascal GPU. This test yielded a dice mean of only 0.87 using U-Net, with each training epoch lasting 11 seconds and a batch size of 16. In our study, we pioneered a new segmentation method that combines batch and group normalization, targeting an improvement in segmentation precision. Our objective is to increase the dice mean and reduce the time required for training.

\subsection{Experimental Results}
\begin{table}[h]
  \centering
  \caption{EVALUATION OF DICE MEAN ACROSS VARIOUS NORMALIZATION TECHNIQUES}\label{tab:dice-normalization}
  \begin{tabular}{cp{20mm}p{20mm}}\toprule Dice Mean & With ELU &With Relu \\\midrule
  Group-Batch Normalization & \textbf{0.95} & {0.94}\\
    Instance-Batch Normalization & \textbf{0.94} & {0.93}\\
    Batch Nomalization & \textbf{0.91} &0.90\\
   
    Layer Normalization &  \textbf{0.89}&0.88 \\

    \bottomrule
  \end{tabular}
\end{table}

For the image segmentation tasks undertaken in this research, we deployed four distinct variants of the U-Net architecture. These variations included U-Net with batch normalization, U-Net with layer normalization, U-Net with batch-group normalization, and U-Net with batch-instance normalization. Each variant of the U-Net architecture was specifically tailored to evaluate the efficacy of different normalization techniques  of image segmentation. This approach was designed to ascertain the impact of various normalization strategies on the segmentation performance. The average dice coefficients are detailed in Table II. Although ELU tends to require more computations compared to ReLU, it mitigates the vanishing gradient problem faced by ReLU for negative values.

Table II presents the dice coefficients from eight distinct experiments, pairing each normalization method with either ELU or ReLU. Our objective was to analyze the impact of these activation functions on segmentation in terms of both accuracy (dice coefficient) and computational speed. Additionally, we aimed to discern the potential enhancements brought by the normalization techniques and identify the most effective method among them.

\begin{table*}[h]
  \centering
  \caption{RESULTS AND ANALYSIS: OUTPUT OF THE MODELS AND EFFICIENCY METRICS EVALUATION }\label{tab:results-analysis}
  \begin{tabular}{cp{23mm}p{23mm}p{23mm}p{23mm}}\toprule & Dice mean & Dice std &Sensitivity &Average perpendicular distance  \\\midrule
  GBU-Net with data augmentation &\textbf {0.97} &\textbf{0.02} &\textbf{0.98} &\textbf{1.88}\\
  GBU-Net without data augmentation &\textbf {0.95} &\textbf{0.02} &\textbf{0.98} &\textbf{1.90}\\
    IBU-Net [21] with data augmentation & {0.96} &{0.02} &{0.98} &{1.91}\\
    IBU-Net  without data augmentation & 0.94 &0.03 &{0.96} &2.02\\
    
    LNU-Net [21] with data augmentation & 0.90 &0.11 & {0.96}&2.29\\
    LNU-Net without data augmentation & 0.89 &0.14 &{0.95}&2.46\\
    
    BNU-Net [18] with data augmentation& 0.93 &0.03 &{0.97} &1.94\\
    BNU-Net  without data augmentation & 0.91 &0.04 &{0.96} &2.06\\
    
    U-Net with data augmentation & 0.88& 0.09 &{0.96} &2.48\\
    U-Net without data augmentation & 0.87&0.11 &{0.95} &2.51\\

    \bottomrule
  \end{tabular}
\end{table*}

%
%\begin{table*}[h]
%  \centering
%  \caption{OUTPUT OF THE LNU-NET AND IBU-NET MODELS AND EFFICIENCY METRICS RESULTS }\label{tab:format-sec}
 % \begin{tabular}{cp{23mm}p{23mm}p{23mm}p{23mm}}\toprule &LNU-Net with data augmentation & LNU-Net without data augmentation &IBU-Net with data augmentation &IBU-Net without data augmentation  \\\midrule
 %   Dice mean & 0.90 &0.89& \textbf{0.96}&0.94\\
 %   Dice std & 0.11 &0.14 &\textbf{0.02}&0.03\\
 %   Sensitivity & 0.97&0.96 &\textbf{0.98} &0.96\\
 %   Average perpendicular distance & 2.46&2.29 &\textbf{1.91} &2.02\\

 %   \bottomrule
 % \end{tabular}
%\end{table*}
%

\begin{figure*}[htbp]
\centering
\includegraphics[width=7.3in,height=4.3in]{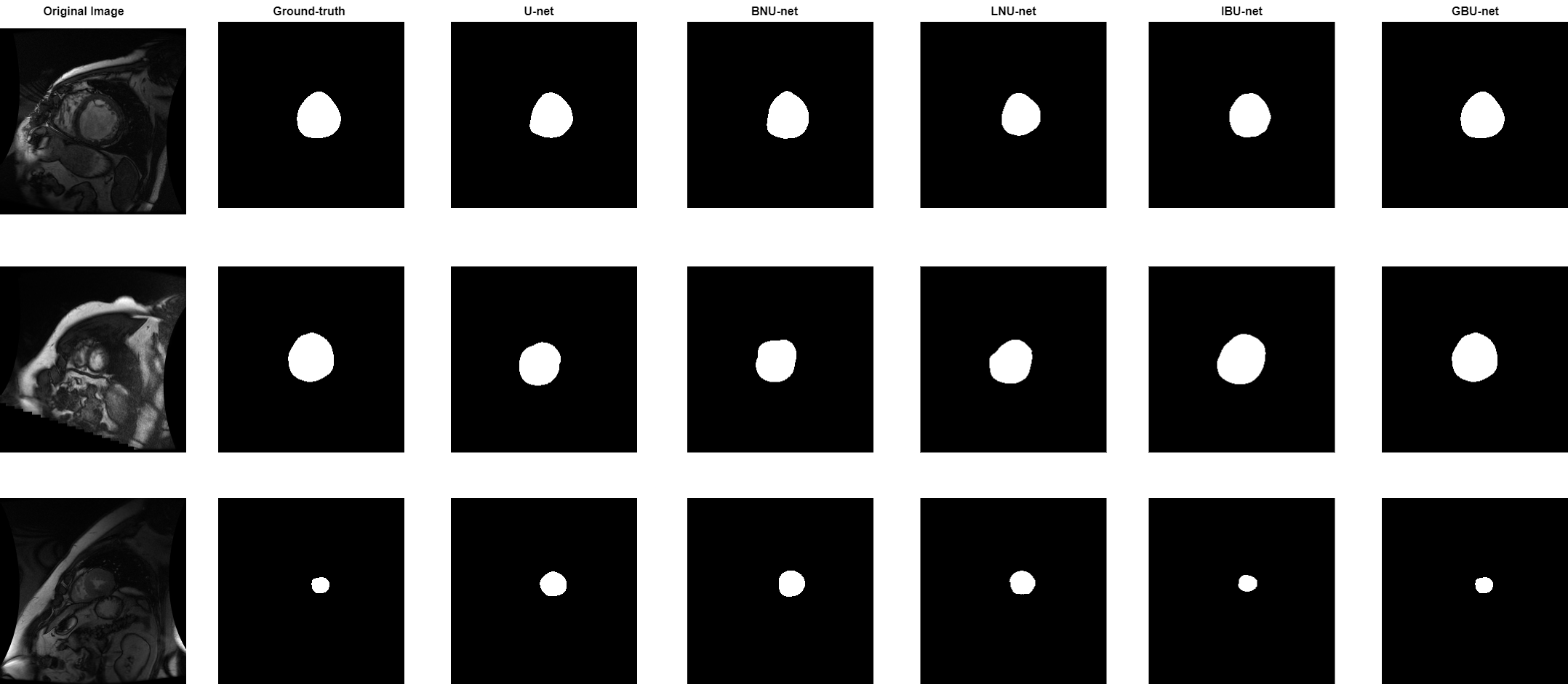}
\caption{Illustrations of segmentation outcomes on unprocessed inputs from three conditions within the Sunnybrook dataset are presented: the first row depicts heart failure with infarction, the second row illustrates hypertrophy, and the third row displays cases from healthy patients.}
\label{fig:segmentation-results}
\end{figure*} 

%\begin{table*}[htbp]
 % \centering
 % \caption{COMPARING THE PERFORMANCE OF U-NET AND BNU-NET ON THE SUNNYBROOK DATASET}\label{tab:format-sec}
%  \begin{tabular}{cp{22mm}p{22mm}p{22mm}p{22mm}}\toprule &U-Net with data augmentation & U-Net without data augmentation &BNU-Net with data augmentation &BNU-Net without data augmentation  \\\midrule
  %  Dice mean & 0.88 &0.87& \textbf{0.93}&0.92\\
  %  Dice std & 0.09 &0.11 &\textbf{0.03}&0.04\\
  %  Sensitivity & 0.96&0.95 &\textbf{0.97} &0.96\\
 % Average perpendicular distance & 2.48&2.51 &\textbf{1.94} &2.06\\
 %   \bottomrule
 % \end{tabular}
%\end{table*}

The experimental data  demonstrate that the integration of the Exponential Linear Unit (ELU) activation function has resulted in a notable enhancement of the average training dice coefficient. In each conducted test, the performance metrics of models utilizing the ELU consistently outperformed those of models driven by the Rectified Linear Unit (ReLU). Consequently, this leads to the conclusion that ELU is more advantageous for tasks involving cardiac MRI image segmentation.

The effectiveness of GBU-Net, as depicted in Figure 4, is comprehensively detailed in Table III. This table provides an insight into the variation in the performance of GBU-Net contingent upon the application or non-application of various data augmentation techniques. Upon analyzing the four normalization techniques detailed in Table III, it becomes clear that batch group normalization is the superior approach for image segmentation tasks, demonstrating the most effective performance among the methods evaluated. 

 When contrasted with the fundamental U-Net architecture, the combined implementation of ELU and batch group normalization significantly increased the dice coefficient by a substantial 8\% – ascending from 0.87 to 0.96.  This escalation signifies a robust and consistent enhancement. Although the acceleration in training velocity wasn't explicitly exhibited in the preceding table, the foundational U-Net model required 11 seconds per epoch during training. However, with the integration of an encoder structure, batch group normalization, and drop-connection, the training time shrunk to 7 seconds per epoch. This constitutes a 36\% surge in speed, translating to the capability to process roughly a third more images within the same time frame. The uptick in speed can be attributed to the encoder design and drop-connection, both of which not only diminish the computational demands during training but also fortify the model against overfitting.

\begin{table}[htbp]
  \centering
 \caption{DICE SCORE AND AVERAGE PERPENDICULAR DISTANCE(APD) OF SEGMENTING THE SUNNYBROOK DATASET,  COMPARED TO THE PERFORMANCE FROM THE STATE OF THE ART METHODS}
   \begin{tabular}{cp{9mm}cp{9mm}cp{9mm}}\toprule & Cases  & Dice Mean & Dice Std & APD (mm)  \\\midrule
  
     \textbf {GBU-Net} &{45} &\textbf {0.97} & \textbf{0.02} & \textbf{1.39}\\\midrule
      Zhou \textit{et al.}[30] &{45} & 0.93 & {0.06} & {-}\\\midrule
    Li \textit{et al.} [20]&{45}  & 0.93 & {0.03} & {1.41}\\\midrule
    Phi [31] & {30} & 0.92 & {0.03} & {1.73} \\\midrule
   X-Y Zhou  [32] &{45} & 0.92 & {-} & {-}\\\midrule
    Queiros \textit{et al.} [33] &{45} & 0.90 & {0.05} & {1.76}\\\midrule
    Medley \textit{et al.} [34] &{45} & 0.90 & {0.04} & {2.00}\\\midrule
    
   Ngo and Carneiro [7] &{45} & 0.90 & {0.03}& {2.08}\\\midrule
   Hu \textit{et al.} [35] &{45} & 0.89 & {0.03} & {2.24}\\\midrule
   Eriksen \textit{et al.} [36] &{45} & 0.89 & {-} & {2.13}\\\midrule
   Huang \textit{et al.} [37]  &{45} &0.89 &0.04 &2.16\\\midrule
   Liu \textit{et al.} [38] &{45} &0.88 &{0.03} &2.36\\\midrule
    Zheng [39]  &{45} &0.88 &{0.07} &2.11\\\bottomrule
\end{tabular}
\label{fig:false-color}
\end{table}

Figure \ref{fig:segmentation-results} exhibits five examples of segmented outputs, each corresponding to a different cardiac condition: heart failure accompanied by infarction, hypertrophy, and those from healthy individuals. To gauge the efficacy of our introduced architectures, we employ the consistent Sunnybrook dataset and identical data partitioning. We then juxtapose our method's performance with other teams that have documented their results on the same 45 patient cases. The magnitude of enhancement in performance differs depending on the chosen normalization approach. As outlined in Table IV, it becomes evident that the utilization of GBU-Net, complemented with data augmentation, outshines other strategies, leading in metrics like dice mean, dice standard deviation, and average perpendicular distance.

The results from the aforementioned algorithm indicates that our proposed method not only increases efficiency but also enhances effectiveness, thus outstripping the performance of traditional deep convolutional networks. This advantage is further reflected in the increased dice coefficient yielded by our approach.

\section{DISCUSSION AND CONCLUSION}In our research, we have enhanced the segmentation of left ventricle cardiac images by improving the U-Net structure. Our preprocessing strategies incorporated batch normalization, layer normalization,  batch-instance normalization, and batch group normalization. Additionally, we employed an encoder architecture and drop-connection to expedite the training process and curb overfitting. 

Our application of advanced normalization techniques significantly improved the dice score for image segmentation. Among the various normalization methods, batch group normalization proved to be the most effective for segmentation tasks. By combining this with the Exponential Linear Unit (ELU), we achieved an exemplary segmentation dice mean of 0.97.

The implementation of our proposed techniques significantly accelerated the training dynamics. Notably, the time required to train each epoch was reduced from 11 seconds to 7 seconds. Compared to the traditional U-Net, our customized variant demonstrated an improved dice score and faster processing times, indicating enhanced results alongside decreased computational requirements.

Given that Cardiac CINE magnetic resonance imaging is deemed the benchmark for evaluating cardiac functionality [40], our aspiration is for this research to have tangible impacts in real-world medical image segmentation. We envision this endeavor assisting both seasoned medical professionals and patients on their journey towards cardiac health. Nonetheless, the performance of these methods may be compromised by the low quality of images in medical image segmentation. Prospective research could focus on enhancing image quality, advancing image processing algorithms, and formulating algorithms capable of providing immediate feedback to refine and optimize the image acquisition procedure. Deep learning methods for automatically segmenting the left ventricle from short-axis cine MR images have achieved notable accuracy. In the future, we can utilize a unique approach that combines 2D occluding contours and 3D shapes through a variational autoencoder (VAE) and a volumetric autoencoder, where occluding contours extracted from random view projections of a 3D shape dataset [42]. 

The introduction of real-time image guidance marks a significant advancement in interventional medicine [41]. In the future, we can integrate magnetic actuation with an appropriate imaging modality, facilitating the tracking of MRbot within the body and enabling the potential for closed-loop servo control along pre-designed trajectories. It is our aspiration that this work will contribute positively to the practical implementation of image segmentation in medical applications. We aim for these contributions to alleviate the burden on both seasoned physicians and convalescing patients in their battle against cardiac diseases.

%\section*{Acknowledgment}
 %This work was supported by the National Science Foundation award CNS-1646566. All opinions, findings, conclusions or recommendations expressed in this work are those of the authors and do not necessarily reflect the views of our sponsors. 


\begin{thebibliography}{00}


\bibitem{c1}Irshad, M., Muhammad, N., Sharif, M \textit{et al.} ``Automatic segmentation of the left ventricle in a cardiac MR short axis image using blind morphological operation," Eur. Phys J, vol. 133, pp.133-148, 2018. [Online]. Available: https://doi.org/10.1140/epjp/i2018-11941-0


\bibitem{c2}Caroline Petitjean, Jean-Nicolas Dacher, ``A review of seg-mentation methods in short axis cardiac MR images," Medical Image Analysis, vol. 15, pp.169–184, 2011.  [Online]. Available: https://www.sciencedirect.com/science/article/pii/S1361841510001349


\bibitem{c3 }K. E. Melkemi, M. Batouche, and S. Foufou, ``A multiagent system approach for image segmentation
using genetic algorithms and extremal optimization heuristics," Pattern Recognition Letters,
vol. 27, no. 11, pp. 1230-1238, Aug. 2006. [Online]. Available: https://www.sciencedirect.com/science/article/abs/pii/S0167865505003363

\bibitem{c4}Y. W. Chen and Y. Q. Chen, ``Adaptive image segmentation using artificial co-evolving tribes,"
International Journal of Pattern Recognition and Artificial Intelligence, vol. 21, no. 7, pp. 1171, 
2007. [Online]. Available: https://www.worldscientific.com/doi/abs/10.1142/S0218001407005922

\bibitem{c5} P. Huang, H. Cao, and S. Luo, ``An artificial ant colonies approach to medical image segmentation,"
Computer Methods and Programs in Biomedicine, vol. 92, no. 3, pp. 267–273, 2008. [Online]. Available:  https://pubmed.ncbi.nlm.nih.gov/18676053/


\bibitem{c6}Jasmine El-Taraboulsi, Claudia P. Cabrera, Caroline Roney, Nay Aung, ``Deep neural network architectures for cardiac image segmentation,"  2023.
[Online]. Available: https://www.sciencedirect.com/science/article/pii/S2667318523000272

\bibitem{c7} {T.A. Ngo and G.Carneiro,} {``Left ventricle segmentation from cardiac MRI combining level set methods with deep belief networks,"} {2013 IEEE International Conference on Image Processing, pp. 695-699, 2013.} [Online]. Available: https://ieeexplore.ieee.org/document/6738143


\bibitem{c8} {Chen C, Qin C, Qiu H, Tarroni G, Duan J, Bai W, Rueckert D,} {``Deep Learning for Cardiac Image Segmentation: A Review,"} {Front Cardiovasc Med,
vol. 7, Issue 25,
2020, } [Online]. Available: https://www.sciencedirect.com/science/article/pii/S1361841514000954

\bibitem{c9} A. Krizhevsky, I. Sutskever, and G. E. Hinton, ``Imagenet classification with deep convolutional
neural networks," In Proceedings of the 25th International Conference on Neural Information
Processing Systems, vol. 1, pp. 1097-1105, Curran Associates Inc., Red Hook, NY, USA, 2012. [Online]. Available: https://dl.acm.org/doi/10.1145/3065386 


\bibitem{c10} G. E. Hinton, `` Learning multiple layers of representation," Trends in Cognitive Sciences, vol.
11, no. 10, pp. 428–434, 2007. [Online]. Available: https://pubmed.ncbi.nlm.nih.gov/17921042/

\bibitem{c11}D. C. Cire¸san, U. Meier, L. M. Gambardella and J. Schmidhuber, ``Deep, big, simple neural
nets for handwritten digit recognition," In Neural Computation, vol. 22, pp. 3207-3220, 2010. [Online]. Available: https://ieeexplore.ieee.org/document/6797043

\bibitem{c12} S. Ren, K. He, R. Girshick, and J. Sun, ``Faster R-cnn: towards real-time object detection
with region proposal networks,"  In Advances in Neural Information Processing Systems, pp. 91–99, 
2015. [Online]. Available: https://arxiv.org/abs/1506.01497

\bibitem{c13} K. Andrej and F. Li, ``Deep visual-semantic alignments for generating image descriptions," In
Proceedings of the IEEE Conference on Computer Vision and Pattern Recognition, pp. 3128–3137,
2015. [Online]. Available: https://api.semanticscholar.org/CorpusID:8517067


\bibitem{c14}Ronneberger, Olaf \textit{et al.} ``U-Net: Convolutional Networks for Biomedical Image Segmentation,” ArXiv abs/1505.04597 (2015). [Online]. Available: https://arxiv.org/abs/1505.04597


\bibitem{c15}Huang, Gao \textit{et al.} ``Densely Connected Convolutional Networks,” {2017 IEEE Conference on Computer Vision and Pattern Recognition (CVPR),  pp. 2261-2269, 2016.} [Online]. Available: https://api.semanticscholar.org/CorpusID:9433631



\bibitem{c16}Zhou Z, Siddiquee MM, Tajbakhsh N, Liang J, ``Unet++: A nested u-net architecture for medical image segmentation,” {Deep Learning in Medical Image Analysis and Multimodal Learning for Clinical Decision Support, pp. 3-11, 2018 Sep 20, Springer, Cham.}  [Online]. Available: https://arxiv.org/abs/1807.10165


\bibitem{c17}Oktay O, Schlemper J, Folgoc LL, Lee M, Heinrich M, Misawa K, Mori K, McDonagh S, Hammerla NY, Kainz B, Glocker B, ``Attention u-net: Learning where to look for the pancreas,” arXiv preprint arXiv:1804.03999, 2018 Apr 11. [Online]. Available: https://arxiv.org/abs/1804.03999


\bibitem{c18}W. Chu \textit{et al.} ``BNU-Net: A Novel Deep Learning Approach for LV MRI Analysis in Short-Axis MRI," 2019 IEEE 19th International Conference on Bioinformatics and Bioengineering (BIBE), Athens, Greece, 2019, pp. 731-736, doi: 10.1109/BIBE.2019.00137. [Online]. Available: https://ieeexplore.ieee.org/document/8941959


\bibitem{c19 }D. Fourure, R. Emonet, E. Fromont, D. Muselet, A. Tremeau, and C. Wolf,``
Residual conv-deconv grid network for semantic segmentation,"  arXiv preprint
arXiv:1707.07958, 2017.  [Online]. Available: https://api.semanticscholar.org/CorpusID:215826069

\bibitem{c20 }Li, Chuchen \textit{et al.} ``Cardiac MRI segmentation with focal loss constrained deep residual networks.” Physics in medicine and biology, vol. 66, 2021. [Online]. Available:  doi:10.1088/1361-6560/ac0bd3


\bibitem{c21} Wenhui Chu and Nikolaos V. Tsekos, ``Two Deep Learning Approaches for Automated Segmentation of Left Ventricle in Cine Cardiac MRI," In Proceedings of the 2022 12th International Conference on Bioscience, Biochemistry and Bioinformatics, pp. 7-13, 2022. [Online]. Available: https://doi.org/10.1145/3510427.3510429

\bibitem{c22} Santurkar, Shibani \textit{et al.} ``How does batch normalization help optimization?" Advances in neural information processing systems, 2018. [Online]. Available: https://arxiv.org/abs/1805.11604


\bibitem{c23}{Jimmy Lei Ba, Jamie Ryan Kiros, Geoffrey E. Hinton,} {``Layer normalization,"} {preprint arXiv: 1607.06450. 2016 Jul 21.} [Online]. Available: https://arxiv.org/abs/1607.06450


\bibitem{c24}Ulyanov D, Vedaldi A, Lempitsky V, ``Instance normalization: The missing ingredient for fast stylization,”  arXiv preprint arXiv:1607.08022, 2016 Jul 27. [Online]. Available: https://arxiv.org/abs/1607.08022


\bibitem{c25} Wu, Y., He, K, ``Group Normalization," International Journal of Computer Vision, pp.742–755, 2020. [Online]. Available: https://doi.org/10.1007/s11263-019-01198-w


\bibitem{c26}Shorten, C., Khoshgoftaar, T.M. ``A survey on Image Data Augmentation for Deep Learning," J Big Data, vol. 6, pp. 60, 2019. [Online]. Available: https://doi.org/10.1186/s40537-019-0197-0 


\bibitem{c27} Zhou, Xiao-Yun, Jiacheng Sun, Nanyang Ye, Xu Lan, qijun luo, Bolin Lai, Pedro M. Esperança, Guang-Zhong Yang and Zhenguo Li, ``Batch Group Normalization,” 2020. [Online]. Available: https://api.semanticscholar.org/CorpusID:227305554


\bibitem{c28} ``Sunnybrook Cardiac Data." [Online]. Available: https://www.cardiacatlas.org/studies/sunnybrook-cardiac-data/


\bibitem{c29}{P. Y. Simard, D. Steinkraus and J. C. Platt,} ``Best practices for convolutional neural networks applied to visual document analysis,"{Seventh International Conference on Document Analysis and Recognition, Proceedings, pp. 958-963, 2003.} [Online]. Available: https://ieeexplore.ieee.org/document/1227801


\bibitem{c30}{Zhou, X., Li, Q., Shen, M., Li, P., Wang, Z., Yang, G,} {``U-Net Training with Instance-Layer Normalization," }{MMMI@MICCAI (2019).} [Online]. Available: https://arxiv.org/abs/1908.08466

\bibitem{c31 }Tran, Phi, ``A Fully Convolutional Neural Network for Cardiac Segmentation in Short-Axis MRI," 2016. [Online]. Available: https://api.semanticscholar.org/CorpusID:21458057


\bibitem{c32}X. Zhou and G. Yang,  ``Normalization  in  Training  U-Net for 2-D Biomedical Semantic Segmentation," {IEEE Robotics and Automation Letters, vol. 4, pp. 1792-1799, 2018.} [Online]. Available: https://arxiv.org/abs/1809.03783


\bibitem{c33 } Queiroz, Cibele, Beilin, Ruth,  Folke, Carl, Lindborg, Regina, ``Farmland Abandonment: Threat or Opportunity for Biodiversity Conservation? A Global Review. Frontiers in Ecology and the Environment," vol. 288, 2014. [Online]. Available: https://esajournals.onlinelibrary.wiley.com/doi/abs/10.1890/120348


\bibitem{c34 }Medley D O, Santiago C and Nascimento J C, ``Segmenting the left ventricle in cardiac in cardiac mri: from handcrafted to deep region
based descriptors,'' IEEE 16th Int. Symp. on Biomedical Imaging (ISBI), 2019. [Online]. Available: https://ieeexplore.ieee.org/document/8759179


\bibitem{c35} {Hu H, Liu H, Gao Z, Huang L,} {``Hybrid segmentation of left ventricle in cardiac MRI using Gaussian-mixture model and region restricted dynamic programming,"} {Magn Reson Imaging, vol. 31, no. 4, pp. 575-584, 2013.} [Online]. Available: https://pubmed.ncbi.nlm.nih.gov/23245907/


\bibitem{c36} Eriksen, Edvarda Regine Winlund,``A machine learning approach to improve consistency in user-driven medical image analysis,'' PHD Thesis University
of Oslo, 2019. [Online]. Available: https://www.duo.uio.no/handle/10852/69754



\bibitem{c37}Huang, S., Liu J., Lee L.C., Venkatesh S., Teo L., Au C., Nowinski W, {``An image-based comprehensive approach for automatic segmentation of left ventricle from cardiac short axis cine MR images,"} Journal of Digital Imaging, 2011. [Online]. Available: https://www.ncbi.nlm.nih.gov/pmc/articles/PMC3138938/ 


\bibitem{c38}Liu, H., Hu H., Xu X., Song E,  {``Automatic left ventricle segmentation in cardiac MRI using topological stable-state thresholding and region restricted dynamic programming,"} Academic Radiology, vol. 19, no. 6, pp. 723-731, 2012. [Online]. Available:
https://doi.org/10.1016/j.acra.2012.02.011



\bibitem{c39 } Zheng Q, ``Deep learning for robust segmentation and explainable analysis of 3d and dynamic cardiac images'', PHD Thesis COMUE
Université Côte d’Azur, 2019. [Online]. Available: https://theses.hal.science/tel-02083415


\bibitem{c40}Küstner, T., Fuin, N., Hammernik, K.\textit{et al.} ``CINENet: deep learning-based 3D cardiac CINE MRI reconstruction with multi-coil complex-valued 4D spatio-temporal convolutions," Sci Rep, vol. 10, 2020. [Online]. Available: https://www.nature.com/articles/s41598-020-70551-8



\bibitem{c41} Wenhui Chu, Khang Tran, and Nikolaos V. Tsekos, ``Simulations of MRI Guided and Powered Ferric Applicators for Tetherless Delivery of Therapeutic Interventions," In Proceedings of the 2022 12th International Conference on Bioscience, Biochemistry and Bioinformatics (ICBBB '22), pp. 29-37, 2022. [Online]. Available: https://doi.org/10.1145/3510427.3510432


\bibitem{c42} Aobo Jin, Qiang Fu, and Zhigang Deng, ``Contour-based 3D Modeling through Joint Embedding of Shapes and Contours," In Symposium on Interactive 3D Graphics and Games (I3D '20). Association for Computing Machinery, New York, NY, USA, Article 9, pp. 1–10, 2020.  [Online]. Available: https://doi.org/10.1145/3384382.3384518







% \bibitem{c3}Stalidis, G., Maglaveras, N., Efstratiadis, S., Dimitriadis, A., Pappas, C. ``Model-based processing scheme for quantitative 4-D cardiac MRI analysis,” IEEE Trans Inf Technol Biomed, 2002, 6 (1), 59–72. [Online]. Available: https://pubmed.ncbi.nlm.nih.gov/11936598/


% \bibitem{c4}R. Poudel, P. Lamata, and G. Montana, “Recurrent fully convolutionalneural networks for multi-slice MRI cardiac segmentation,” arXivpreprint arXiv:1608.03974, 2016 [Online]. Available: https://arxiv.org/abs/1608.03974

% \bibitem{c5}Jasmine El-Taraboulsi, Claudia P. Cabrera, Caroline Roney, Nay Aung,, ``Deep neural network architectures for cardiac image segmentation,"  2023.
% [Online]. Available: https://www.sciencedirect.com/science/article/pii/S2667318523000272

% \bibitem{c6} {T.A. Ngo and G.Carneiro,} {``Left ventricle segmentation from cardiac MRI combining level set methods with deep belief networks,"} {2013 IEEE International Conference on Image Processing, 2013, pp. 695-699, doi: 10.1109/ICIP.2013.6738143.} [Online]. Available: https://ieeexplore.ieee.org/document/6738143


% \bibitem{c7} {Chen C, Qin C, Qiu H, Tarroni G, Duan J, Bai W, Rueckert D,} {``Deep Learning for Cardiac Image Segmentation: A Review,"} {Front Cardiovasc Med,
% Volume 7, Issue 25,
% 2020, } [Online]. Available: https://www.sciencedirect.com/science/article/pii/S1361841514000954

% \bibitem{c8} {Hu H, Liu H, Gao Z, Huang L," } {``Hybrid segmentation of left ventricle in cardiac MRI using Gaussian-mixture model and region restricted dynamic programming,"} {Magn Reson Imaging. 2013 May; 31(4):575-84. doi: 10.1016/j.mri.2012.10.004. Epub 2012 Dec 14. PMID: 23245907.} [Online]. Available: https://pubmed.ncbi.nlm.nih.gov/23245907/



% \bibitem{c9}Huang, S., Liu J., Lee L.C., Venkatesh S., Teo L., Au C., Nowinski W. 2011. {"An image-based comprehensive approach for automatic segmentation of left ventricle from cardiac short axis cine MR images,"} Journal of Digital Imaging 24 598608. [Online]. Available: https://www.ncbi.nlm.nih.gov/pmc/articles/PMC3138938/ 

% \bibitem{c10}Margeta, J., Geremia E., Criminisi A., Ayache N. 2012. {``Layered spatio-temporal forests for left ventricle segmentation from 4D cardiac MRI data,"} {MICCAI workshop: Statistical Atlases and Computational Models of the Heart (STACOM).} [Online]. Available: https://www.microsoft.com/en-us/research/publication/layered-spatio-temporal-forests-for-left-ventricle-segmentation-from-4d-cardiac-mri-data/


% \bibitem{c11}Liu, H., Hu H., Xu X., Song E. 2012. {``Automatic left ventricle segmentation in cardiac MRI using topological stable-state thresholding and region restricted dynamic programming,"} Academic Radiology. [Online]. Available:
% https://doi.org/10.1016/j.acra.2012.02.011


% \bibitem{c12}Ronneberger, Olaf et al. ``U-Net: Convolutional Networks for Biomedical Image Segmentation,” ArXiv abs/1505.04597 (2015). [Online]. Available: https://arxiv.org/abs/1505.04597

% \bibitem{c13}Zhou Z, Siddiquee MM, Tajbakhsh N, Liang J. ``Unet++: A nested u-net architecture for medical image segmentation,” {Deep Learning in Medical Image Analysis and Multimodal Learning for Clinical Decision Support 2018 Sep 20 (pp. 3-11). Springer, Cham.}  [Online]. Available: https://arxiv.org/abs/1807.10165

% \bibitem{c14}Oktay O, Schlemper J, Folgoc LL, Lee M, Heinrich M, Misawa K, Mori K, McDonagh S, Hammerla NY, Kainz B, Glocker B. ``Attention u-net: Learning where to look for the pancreas,” arXiv preprint arXiv:1804.03999. 2018 Apr 11. [Online]. Available: https://arxiv.org/abs/1804.03999


% \bibitem{c15} Santurkar, Shibani, et al. ``How does batch normalization help optimization?" Advances in neural information processing systems 31 2018. [Online]. Available: https://arxiv.org/abs/1805.11604


% \bibitem{c16}{Jimmy Lei Ba, Jamie Ryan Kiros, Geoffrey E. Hinton.} {``Layer normalization,"} {preprint arXiv: 1607.06450. 2016 Jul 21.} [Online]. Available: https://arxiv.org/abs/1607.06450



% \bibitem{c17}Nam H, Kim HE. ``Batch-instance normalization for adaptively style-invariant neural networks,” In Advances in Neural Information Processing Systems 2018 (pp. 2558-2567). [Online]. Available:
% https://papers.nips.cc/paper/7522-batch-instance-normalization-for-adaptively-style-invariant-neural-networks.pdf


% \bibitem{c18}Ulyanov D, Vedaldi A, Lempitsky V. ``Instance normalization: The missing ingredient for fast stylization,”  arXiv preprint arXiv:1607.08022. 2016 Jul 27. [Online]. Available: https://arxiv.org/abs/1607.08022

% \bibitem{c19} ``Sunnybrook Cardiac Data." [Online]. Available: https://www.cardiacatlas.org/studies/sunnybrook-cardiac-data/

% \bibitem{c20} E. Shelhamer, J. Long and T. Darrell, ``Fully Convolutional Networks for Semantic Segmentation," {IEEE Transactions on Pattern Analysis and Machine Intelligence, vol. 39, no. 4, pp. 640-651, 1 April 2017, doi: 10.1109/TPAMI.2016.2572683.} [Online]. Available: https://arxiv.org/abs/1411.4038

% \bibitem{c21}X.Zhou and G.Yang,  ``Normalization  in  Training  U-Net for 2-D Biomedical Semantic Segmentation," {IEEE Robotics and Automation Letters 4 (2019): 1792-1799} [Online]. Available: https://arxiv.org/abs/1809.03783

% \bibitem{c22}W. Chu et al., ``BNU-Net: A Novel Deep Learning Approach for LV MRI Analysis in Short-Axis MRI," 2019 IEEE 19th International Conference on Bioinformatics and Bioengineering (BIBE), Athens, Greece, 2019, pp. 731-736, doi: 10.1109/BIBE.2019.00137. [Online]. Available: https://ieeexplore.ieee.org/document/8941959

% \bibitem{c23}{P. Y. Simard, D. Steinkraus and J. C. Platt.} ``Best practices for convolutional neural networks applied to visual document analysis,"{Seventh International Conference on Document Analysis and Recognition, Proceedings, 2003, pp. 958-963} [Online]. Available: https://ieeexplore.ieee.org/document/1227801

% \bibitem{c24}{Zhou, X., Li, Q., Shen, M., Li, P., Wang, Z., Yang, G.} {``U-Net Training with Instance-Layer Normalization," }{MMMI@MICCAI (2019).} [Online]. Available: https://arxiv.org/abs/1908.08466

% \bibitem{c25}Küstner, T., Fuin, N., Hammernik, K. et al. ``CINENet: deep learning-based 3D cardiac CINE MRI reconstruction with multi-coil complex-valued 4D spatio-temporal convolutions," Sci Rep 10, 13710 (2020). https://doi.org/10.1038/s41598-020-70551-8, [Online]. Available: https://www.nature.com/articles/s41598-020-70551-8

% \bibitem{c26}Shorten, C., Khoshgoftaar, T.M. ``A survey on Image Data Augmentation for Deep Learning," J Big Data, 6, 60 (2019). [Online]. Available: https://doi.org/10.1186/s40537-019-0197-0 


% \bibitem{c27}Huang, Gao et al. ``Densely Connected Convolutional Networks,” {2017 IEEE Conference on Computer Vision and Pattern Recognition (CVPR) (2016): pp. 2261-2269.} [Online]. Available: https://api.semanticscholar.org/CorpusID:9433631

% \bibitem{c28 }D. Fourure, R. Emonet, E. Fromont, D. Muselet, A. Tremeau, and C. Wolf.``
% Residual conv-deconv grid network for semantic segmentation," 2007, arXiv preprint
% arXiv:1707.07958, 2017.  [Online]. Available: https://api.semanticscholar.org/CorpusID:215826069

% \bibitem{c29 }Li, Chuchen et al. ``Cardiac MRI segmentation with focal loss constrained deep residual networks.” Physics in medicine and biology vol. 66,13 10.1088/1361-6560/ac0bd3. 2021, doi:10.1088/1361-6560/ac0bd3 [Online]. Available: https://pubmed.ncbi.nlm.nih.gov/34134101/

% \bibitem{c30 }Tran, Phi. ``A Fully Convolutional Neural Network for Cardiac Segmentation in Short-Axis MRI. "(2016).[Online]. Available: https://api.semanticscholar.org/CorpusID:21458057

% \bibitem{c31 } Queiroz, Cibele, Beilin, Ruth,  Folke, Carl, Lindborg, Regina. ``Farmland Abandonment: Threat or Opportunity for Biodiversity Conservation? A Global Review. Frontiers in Ecology and the Environment." (2014), 288. 10.1890/120348. [Online]. Available: https://esajournals.onlinelibrary.wiley.com/doi/abs/10.1890/120348
% \bibitem{c32 }Medley D O, Santiago C and Nascimento J C. ``Segmenting the left ventricle in cardiac in cardiac mri: from handcrafted to deep region
% based descriptors'', IEEE 16th Int. Symp. on Biomedical Imaging (ISBI), 2019. [Online]. Available: https://ieeexplore.ieee.org/document/8759179
% \bibitem{c33 } Eriksen, Edvarda Regine Winlund,``A machine learning approach to improve consistency in user-driven medical image analysis'', PHD Thesis University
% of Oslo, 2019. [Online]. Available: https://www.duo.uio.no/handle/10852/69754

% \bibitem{c34 } Zheng Q. ``Deep learning for robust segmentation and explainable analysis of 3d and dynamic cardiac images'', PHD Thesis COMUE
% Université Côte d’Azur, 2019. [Online]. Available: https://theses.hal.science/tel-02083415

% \bibitem{c35 }K. E. Melkemi, M. Batouche, and S. Foufou. ``A multiagent system approach for image segmentation
% using genetic algorithms and extremal optimization heuristics," Pattern Recognition Letters,
% vol. 27, no. 11, pp. 1230-1238, Aug. 2006. [Online]. Available: https://www.sciencedirect.com/science/article/abs/pii/S0167865505003363


% \bibitem{c36}Y. W. Chen and Y. Q. Chen. ``Adaptive image segmentation using artificial co-evolving tribes,"
% International Journal of Pattern Recognition and Artificial Intelligence, vol. 21, no. 7, pp. 1171, 
% 2007. [Online]. Available: https://www.worldscientific.com/doi/abs/10.1142/S0218001407005922

% \bibitem{c37} P. Huang, H. Cao, and S. Luo. ``An artificial ant colonies approach to medical image segmentation,"
% Computer Methods and Programs in Biomedicine, vol. 92, no. 3, pp. 267–273, 2008. [Online]. Available:  https://pubmed.ncbi.nlm.nih.gov/18676053/

% \bibitem{c38} A. Krizhevsky, I. Sutskever, and G. E. Hinton. ``Imagenet classification with deep convolutional
% neural networks," In Proceedings of the 25th International Conference on Neural Information
% Processing Systems, vol. 1, pp. 1097-1105, Curran Associates Inc., Red Hook, NY, USA, 2012. [Online]. Available: https://dl.acm.org/doi/10.1145/3065386 


% \bibitem{c39} G. E. Hinton. `` Learning multiple layers of representation," Trends in Cognitive Sciences, vol.
% 11, no. 10, pp. 428–434, 2007. [Online]. Available: https://pubmed.ncbi.nlm.nih.gov/17921042/

% \bibitem{c40}D. C. Cire¸san, U. Meier, L. M. Gambardella and J. Schmidhuber. ``Deep, big, simple neural
% nets for handwritten digit recognition," In Neural Computation, vol. 22, pp. 3207-3220, 2010. [Online]. Available: https://ieeexplore.ieee.org/document/6797043

% \bibitem{c41} S. Ren, K. He, R. Girshick, and J. Sun. ``Faster R-cnn: towards real-time object detection
% with region proposal networks,"  In Advances in Neural Information Processing Systems, pp. 91–99, 
% 2015. [Online]. Available: https://arxiv.org/abs/1506.01497

% \bibitem{c42} K. Andrej and F. Li. ``Deep visual-semantic alignments for generating image descriptions," In
% Proceedings of the IEEE Conference on Computer Vision and Pattern Recognition, pp. 3128–3137,
% 2015. [Online]. Available: https://api.semanticscholar.org/CorpusID:8517067


% \bibitem{c43} Wenhui Chu and Nikolaos V. Tsekos. ``Two Deep Learning Approaches for Automated Segmentation of Left Ventricle in Cine Cardiac MRI," In Proceedings of the 2022 12th International Conference on Bioscience, Biochemistry and Bioinformatics, New York, NY, USA, 2022 [Online]. Available: https://doi.org/10.1145/3510427.3510429



% \bibitem{c44} Wu, Y., He, K. ``Group Normalization," International Journal of Computer Vision, pp.742–755 (2020). [Online]. Available: https://doi.org/10.1007/s11263-019-01198-w


% \bibitem{c45} Zhou, Xiao-Yun, Jiacheng Sun, Nanyang Ye, Xu Lan, qijun luo, Bolin Lai, Pedro M. Esperança, Guang-Zhong Yang and Zhenguo Li. ``Batch Group Normalization,” (2020). [Online]. Available: https://api.semanticscholar.org/CorpusID:227305554
\end{thebibliography}
\end{document}